\documentclass[runningheads]{llncs}

\usepackage{mytitlesec}
\titlespacing\section{0pt}{12pt plus 4pt minus 2pt}{5pt plus 2pt minus 2pt}
\titlespacing\subsection{0pt}{12pt plus 4pt minus 2pt}{5pt plus 2pt minus 2pt}
\titlespacing\subsubsection{0pt}{12pt plus 4pt minus 2pt}{5pt plus 2pt minus 2pt}

\usepackage{eccv}

\usepackage{eccvabbrv}

\usepackage{graphicx}
\usepackage{booktabs}

\usepackage[accsupp]{axessibility}  

\usepackage[pagebackref,breaklinks,colorlinks,citecolor=eccvblue]{hyperref}
\usepackage[dvipsnames]{xcolor}

\usepackage{multirow}
\usepackage{algorithm}
\usepackage{algorithmic}
\usepackage[algo2e,ruled,vlined,boxed,commentsnumbered]{algorithm2e}
\usepackage{threeparttable}

\usepackage{soul}
\usepackage{colortbl}

\SetKwInput{KwInput}{Input}                
\SetKwInput{KwOutput}{Output}              

\def\minus#1{{($- #1$)}}

\usepackage{orcidlink}

\begin{document}

\title{ActiveAD: Planning-Oriented Active Learning for End-to-End Autonomous Driving} 

\titlerunning{Planning-Oriented Active Learning for End-to-End Autonomous Driving}

\author{Han Lu$^{1,*}$,
Xiaosong Jia$^{1,*}$,
Yichen Xie$^{2}$, Wenlong Liao$^{1}$, Xiaokang Yang$^{1}$, Junchi Yan$^{1,\dag}$ 
}

\authorrunning{Lu et al.}


\institute{$^{1}$ Department of Computer Science and Engineering, Shanghai Jiao Tong University \\ $^{2}$ University of California, Berkeley
\\
$^*$ Equal contributions \quad $^\dag$ Correspondence author}

\maketitle

\begin{abstract}


End-to-end differentiable learning for autonomous driving (AD) has recently become a prominent paradigm. One main bottleneck lies in its voracious appetite for high-quality labeled data \textit{e.g.} 3D bounding boxes and semantic segmentation, which are notoriously expensive to manually annotate. The difficulty is further pronounced due to the prominent fact that the behaviors within samples in AD often suffer from long tailed distribution. In other words, a large part of collected data can be trivial (e.g. simply driving forward in a straight road) and only a few cases are safety-critical. In this paper, we explore a practically important yet under-explored problem about how to achieve sample and label efficiency for end-to-end AD. Specifically, we design a planning-oriented active learning method which progressively annotates part of collected raw data according to the proposed diversity and usefulness criteria for planning routes. Empirically, we show that our planning-oriented approach could outperform general active learning methods by a large margin. \textbf{Notably, our method achieves comparable performance with state-of-the-art end-to-end AD methods - by using only 30\% nuScenes data}. We hope our work could inspire future works to explore end-to-end AD from a data-centric perspective in addition to methodology efforts.


\end{abstract}    
\section{Introduction}

Autonomous driving (AD), as one of the most exciting applications of AI, has drawn increasing attention. Traditional AD systems are usually module-based which divide the driving task into sub-tasks: perception~\cite{li2022bevformer,huang2021bevdet,liu2023bevfusion}, prediction~\cite{shi2022motion,jia2022multi,jia2022temporal,jia2023hdgt}, planning~\cite{treiber2000congested,Dauner2023CORL}, etc. However, modular systems suffer from error accumulations, less principled optimization, and redundant computations due to the separate training objectives of each sub-task, which limit the performance upper bound of these systems~\cite{chen2023end}. On the other hand, the success of LLM~\cite{brown2020language,openai2023gpt4} has demonstrates the power of the data-driven scalable paradigm~\cite{wu2023policy,yang2023llm4drive}. Aiming to overcome the drawbacks of the traditional module-based systems and embrace the power of data, end-to-end AD (E2E-AD) becomes a hot topic recently~\cite{hu2023planning}. 

\begin{figure}[t!]
    \centering     
    \includegraphics[width=\linewidth]{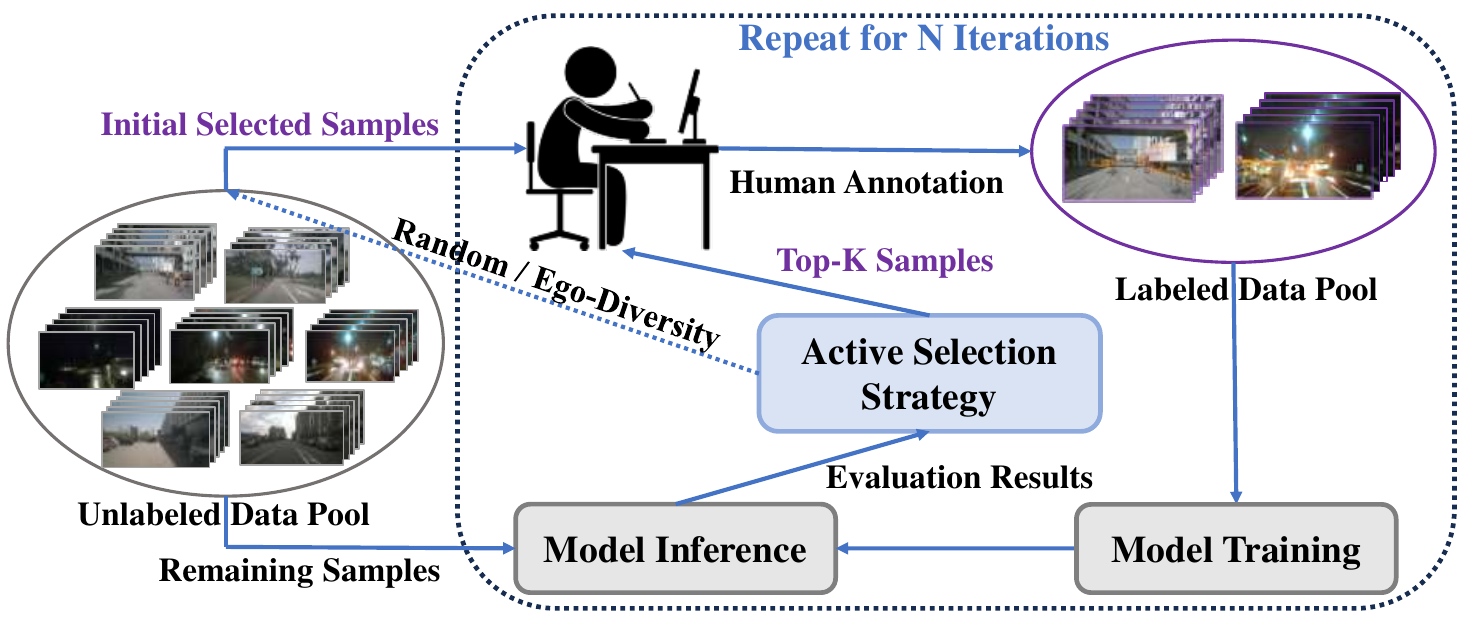}    \caption{\textbf{Active Learning scheme for end-to-end Autonomous Driving.} We formulate the pipeline and meticulously design the task-specific Active Selection Strategy for selecting initial samples and the incremental samples in the following iterations.}  
    \label{fig:Active_Pipeline}
    \vspace{-10pt}
\end{figure}
One key success factor of LLM is the huge amount of nearly free texts available on the Internet, which is not the case in AD where state-of-the-art E2E-AD systems such as UniAD~\cite{hu2023planning} and VAD~\cite{jiang2023vad} are still confined by supervised learning, which require fine-grained annotations including 3D bounding boxes of agents and semantic segmentation for lanes and traffic signs. Such annotations are expensive and as a result, \textbf{labeling becomes one bottleneck of the scaling up process of these end-to-end methods}. Even worse, it is widely acknowledged in the community that the AD task has serious long-tailed issues~\cite{jain2021autonomy}. In other words, a large part of collected data is trivial \textit{e.g.} simply driving forward in a straight road, and only a few cases are safety-critical, which further makes the data-driven methods less applicable.

To deal with the above issues, in this work, we first investigate the seminal question that: \emph{Do we really need to annotate all collected raw data to achieve best performance}? By empirical studies, we demonstrate that the answer is \emph{NO}. Further, we explore the way to select the most useful samples to annotate for planning , which belongs to the task of active learning~\cite{zhan2022comparative}. Different from existing literature focusing on the perception part~\cite{luo2023exploring}, inspired by the planning-oriented philosophy in UniAD~\cite{hu2023planning}, we design our active learning method, called ActiveAD, based on the planning route and scores, to directly optimize planning.

There are several major gaps for adopting existing active learning methods~\cite{gal2017deep,kirsch2019batchbald,ash2019deep,yoo2019learning,sinha2019variational} to AD. On one hand,  data in AD often involves rich multi-modality information such as video streams, driving trajectories, and miscellaneous meta information like vehicle speed while existing general active learning methods usually only consider a single-modal image as input. On the other hand, AD tasks could be complex beyond classification, to which however existing works are often confined~\cite{gal2017deep,kirsch2019batchbald,ash2019deep}. It calls for adaption to better handle the information among diverse inputs and optimization targets.


 
Fig.~\ref{fig:Active_Pipeline} shows the designed scheme of active learning paradigm for end-to-end AD and we aim to overcome these challenges and better utilize task-relevant information. 
In the initial sample selection stage, ActiveAD introduces Ego-Diversity as a replacement for the commonly used random selection in traditional AD paradigms~\cite{sener2018active, sinha2019variational}. Ego-Diversity effectively utilizes some nearly free information inside raw AD data by considering factors including weather, lighting, and vehicle speed. During the iterative process of active sample selection, we propose three intuitive and effective metrics: Displacement Error, Soft Collision, and Agent Uncertainty. The Displacement Error effectively utilizes the recorded ego trajectory and serves as a concise yet essential metric. Soft Collision computes the potential of collisions based on the predicted trajectory of the ego vehicle and the trajectories of other objects, which is a continuous version of collision rate. 
Agent Uncertainty focuses on complex road condition, which assesses the uncertainty of other vehicles.
 
Extensive experiments are conducted to validate the proposed  ActiveAD. It significantly outperforms general active learning methods. Under the $30\%$ annotation budget, ActiveAD also achieves comparable or even slightly better planning performance than state-of-the-art methods trained on the complete dataset. In the ablation study, we conduct a detailed analysis of the contribution and effectiveness of using the designed metrics, examining the robustness of performance across different scenarios. We also provide visualizations and analyses of the results for different selection choices. The contributions include:


\begin{itemize}
    \item To the best of our knowledge, \textbf{we are the first to delve into the data problems of E2E-AD.} We also give a simple yet effective solution to identify and annotate the valuable data for planning within a limited budget.
    \item  Based on the planning-oriented philosophy of end-to-end methods, we design the \textbf{novel task-specific diversity and uncertainty measurement} for the planning routes.
    \item Extensive experiments and ablation studies demonstrate the effectiveness of our approach. \textbf{ActiveAD outperforms the general peer methods by a large margin and achieves comparable performance with SOTA method with complete labels, using only $30\%$ nuScenes data}.
    
\end{itemize}

\section{Related Work}

\subsection{End-to-End Autonomous Driving}
The concept of end-to-end autonomous driving has roots dating back to the 1980s~\cite{pomerleau1988alvinn}. In the era of deep learning, early efforts focused on the straightforward mapping~\cite{muller2005off}. 
 Subsequently, \cite{zhang2021roach,li2024think2drive} explored the application of reinforcement learning to develop an end-to-end driving policy. Some state-of-the-art student models~\cite{wu2022trajectoryguided,hu2022model} are developed based on them while PlanT~\cite{Renz2022CORL} suggested employing a Transformer for the teacher model. LBC~\cite{chen2020learning} and DriveAdapter~\cite{jia2023driveadapter} involved initially training a teacher model with privileged inputs. In later works, multiple sensors are used. 
 Transfuser~\cite{Prakash2021CVPR,Chitta2022PAMI} employed a Transformer for camera and LiDAR fusion. LAV~\cite{chen2022lav} adopted PointPainting~\cite{vora2020pointpainting}. Interfuser~\cite{shao2022interfuser} injected safety-enhanced rules during the decision-making process. 
 ThinkTwice~\cite{jia2023thinktwice} introduced a DETR-like scalable decoder paradigm for the student model. 
 ReasonNet proposed specific modules for student models to better exploit temporal and global information. In \cite{Jaeger2023ICCV}, they suggested formulating the output of the student as classification problems to avoid averaging. 
ST-P3~\cite{hu2022st} unified the detection, prediction, and planning tasks into the form of BEV segmentation. UniAD~\cite{hu2023planning} adopted Transformer to connect different tasks. Further, VAD~\cite{jiang2023vad} reduced some potential redundant modules in UniAD while demonstrating better performance.

\subsection{Active Learning}
\label{sec:activelearning}
Active learning algorithms exploit the limited annotation budget by selecting the most informative samples for labeling. They select data samples based on the criterion of either uncertainty or diversity. Uncertainty-based algorithms prefer those difficult samples most confusing for the models. The difficulty of each data sample may be measured by prediction entropy~\cite{joshi2009multi,luo2013latent}, prediction inconsistency~\cite{gao2020consistency}, loss estimation~\cite{yoo2019learning} or its potential influence for model training~\cite{freytag2014selecting,liu2021influence}. Alternatively, other methods pay attention to the diversity of the selected subset. Some early work~\cite{sener2018active,sinha2019variational} mainly considers the representation diversity in the global image level, while following papers~\cite{agarwal2020contextual,liang2022exploring} dig into the regional information to deal with fine-grained detection or segmentation tasks.  Furthermore, some recent work~\cite{xie2023towards,xie2023active,yi2022pt4al} utilizes the strong representation ability of models pretrained on large datasets to measure the image diversity of the target dataset more accurately. Recently, CRB~\cite{luo2023exploring} has pioneered the extension of active learning to LiDAR-based 3D object detection in autonomous driving.

However, most prior works focus on the traditional tasks like classification, detection, or segmentation, but the recently prominent planning-oriented end-to-end AD setting is hardly explored. Instead of just simple prediction probability, The task model outputs the future ego-vehicle trajectory. Besides, this task requires to reason from the interaction~\cite{jia2021ide} between ego-vehicle and surroundings, which cannot be reflected from superficial visual patterns. To this end, we fill in this gap by devising novel uncertainty and diversity metrics for active learning of end-to-end AD.

\section{Formulation of Active Learning for AD}
State-of-the-art end-to-end AD methods~\cite{hu2022goal,jiang2023vad} usually take raw sensor data as inputs and generate the planned trajectories for the ego vehicle. To facilitate training and mitigate overfitting, additional annotations like 3D bounding boxes of agents and semantic segmentation of lanes~\cite{caesar2020nuscenes} are used. Since the collected raw data is typically in the form of clips containing multiple temporal frames of surrounding images and canbus information, organizing the annotations at the clip-level offers several benefits. Firstly, it streamlines the annotation process by providing a coherent context for labeling. Secondly, it enables the establishment of spatiotemporal connections between objects. Therefore, we choose to treat each clip as a distinct unit, rather than considering individual frames. This is also in accordance with practice in AD research~\cite{caesar2020nuscenes}.

Formally, we define the active learning task for end-to-end AD as follows: denote $\mathcal{I}^t$ as the raw sensor data in the frame $t$ where $t \in [T]=\{1, 2, ..., T\}$ and $T$ is the length of its corresponding clip $\mathcal{S}_i$. Apart from the raw sensor data, the recorded trajectory $\tau_i$ and states $e_i$ (speed $v_i$ and driving commands $\textit{cmd}_i$) of the ego vehicle, weather condition $w_i$ (Sunny or Rainy) and the lighting condition $l_i$ (Day or Night) are also annotation-free or extremely cheap to obtain. For simplicity, we denote these easy-to-obtain labels as $\mathcal{O}_i = (e_i, w_i, l_i)$. For the scene that has not been meticulously annotated (e.g., without annotations of 3D bounding boxes and semantic segmentation), we can represent such information as $X_i = (\mathcal{S}_i, \tau_i, \mathcal{O}_i)$ where $i \in [N]$ and $N$ is the number of scenes.

For the labels that require meticulous annotation, we denote them as $Y_i$. $Y_i = (\mathcal{A}_i, \mathcal{B}_i, \mathcal{C}_i)$ where $\mathcal{A}_i$ donates attributes (visibility, activity, and pose), $\mathcal{B}_i$ denotes the 3D bounding box 
and $\mathcal{C}_i$ donates the semantic segmentation of lanes~\cite{caesar2020nuscenes}.

Initially, we have the access to the unlabeled data pool $\mathcal{P}^u = \{X_i\}_{i \in [N]}$. Under the given annotation budget $B$ where $|B| < N$, one should select the index set $\mathcal{K} = \{k_i \in [N]\}_{i \in [B]}$ to obtain the subset $\mathcal{P}^u_{\mathcal{K}} = \{X_{k_i}\}_{i \in [B]} \subset \mathcal{P}^u$ from $\mathcal{P}^u$ and acquire the related labels $\{Y_{k_i}\}_{i \in [B]}$. Then the models are trained on the labeled set $\mathcal{P}^l_\mathcal{K} = \{(X_{k_i}, Y_{k_i})\}_{i \in [B]}$. The objective is to choose the sampling strategy to select the labeled set under the budget to minimize the expectation error of the model, which usually refers to the L2 loss and collision ratio~\cite{hu2022goal,jiang2023vad} in end-to-end AD.

The active selection process involves the following steps: 1) Select a subset of data as the initial set. 2) Train a model based on the current data. 3) Utilize the trained model's features and outputs to select a new subset of data based on a designed strategy. 4) Repeat steps 2 and 3 until the budget is reached. Fig.~\ref{fig:Active_Pipeline} demonstrates the pipeline and the process combined with our method is detailed in Sec.~\ref{sec:pipeline}.

\section{The Proposed ActiveAD}
\label{sec:active}

We provide a detailed description of our method ActiveAD, within the framework of end-to-end AD. Leveraging the characteristics of data specific to AD, we devise corresponding metrics for diversity and uncertainty. Sec.~\ref{sec:initSelection} introduces the methodology for designing diversity metrics, which are utilized as criteria for selecting the initial set. Sec.~\ref{sec:increSelection} presents the design of uncertainty metrics to identify more challenging data samples. Sec.~\ref{sec:pipeline} summarizes the entire ActiveAD process and provides a detailed algorithmic depiction.

\subsection{Initial Sample Selection for Labeling}
\label{sec:initSelection}
For active learning in computer vision, the initial sample selection is often solely based on the raw images  without extra information or learned features, leading to the common practice of \textbf{Random} initialization~\cite{sener2018active,sinha2019variational,yoo2019learning,kim2021task,parvaneh2022active}. For AD, there is additional prior information to leverage. Specifically, when collecting data from sensors, one can simultaneously record conventional information such as the speed and trajectory of the ego vehicle. Additionally, weather and lighting conditions are generally continuous and easy to annotate in the clip-level. These information can benefit making informed choices for the initial set selection. Therefore, we design the \textbf{Ego-Diversity} metric for initial selection.

\textbf{Ego-Diversity} consists of three components: 1) weather-lighting 2) driving commands 3) average speed. Inspired by the setting in \cite{liu2023bevfusion, zhu2023understanding}, we firstly divide the complete dataset into four mutually exclusive subsets: Day-Sunny (DS), Day-Rainy (DR), Night-Sunny (NS), Night-Rainy (NR), using the description in nuScenes~\cite{caesar2020nuscenes}. Secondly, We categorize each subset based on the number of left, right, and straight driving commands~\cite{hu2022st, hu2022goal, jiang2023vad} within a complete clip into four categories: Turn Left (L), Turn Right (R), Overtake (O), Go Straight (S). We design a threshold $\tau_{c}$, where if the numbers of left and right commands in a clip are both greater than or equal to the threshold $\tau_{c}$, we consider it as an overtaking behavior in this clip. If only the number of left commands is greater than the threshold $\tau_{c}$, it indicates a left turn. If only the number of right commands is greater than the threshold $\tau_{c}$, it indicates a right turn. All the other cases are considered as going straight. Thirdly, we calculate the average speed in each scene and sort them in ascending order in the related subset. 

\begin{figure}[tb!]
    \centering 
    \includegraphics[width=0.9\linewidth]{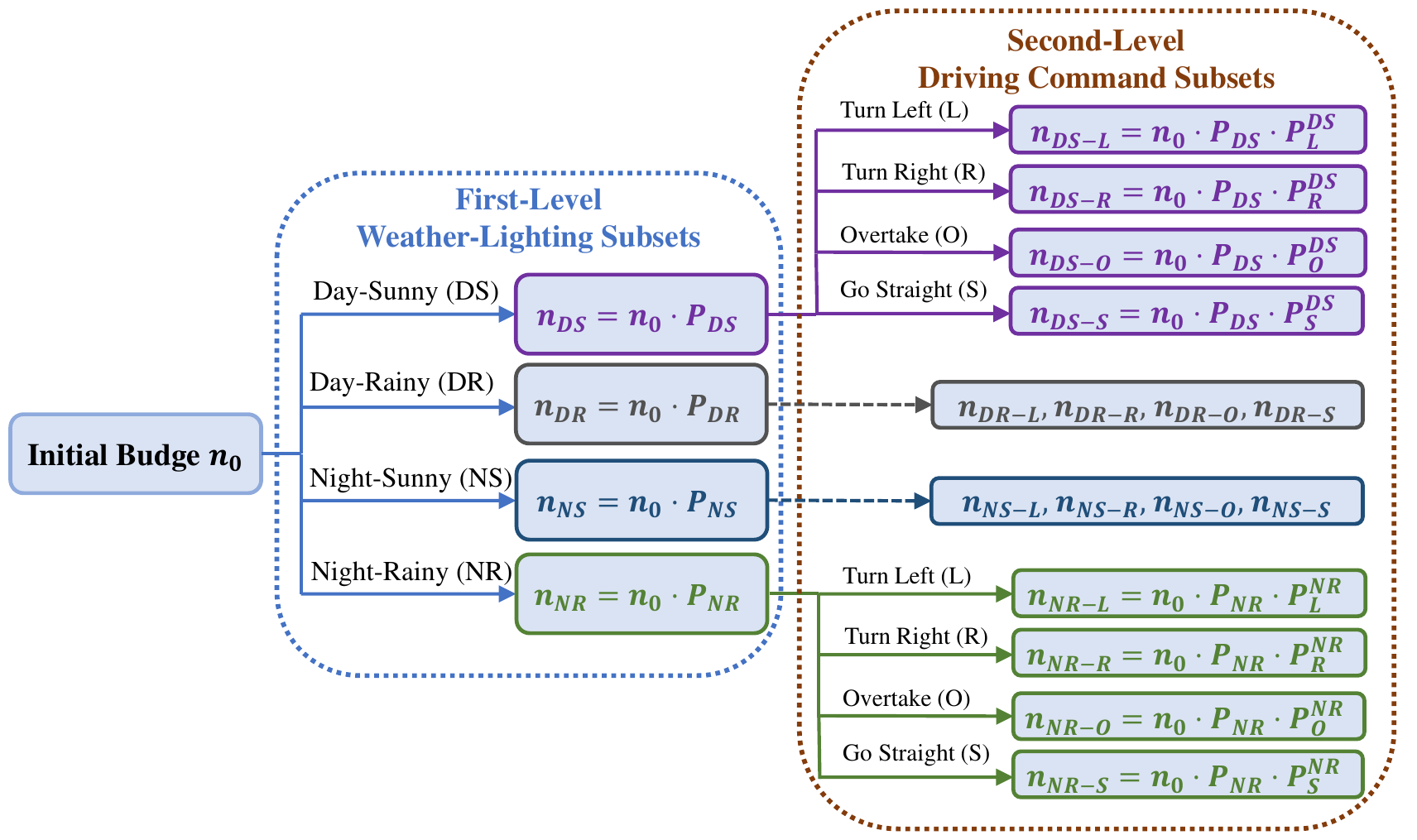}
    \caption{\textbf{Ego-Diversity Initialization.} The process of allocating the given budget to various subsets is illustrated. Firstly, the first-level subsets are divided based on the proportion of Weather-Lighting conditions. Then, the second-level subsets are formed based on the respective proportions of Driving Commands. Finally, within each subset of the Second-level stage, samples are sorted based on speed and uniformly selected. 
    }  
    \vspace{-10pt}
    \label{fig:egodiversity}
\end{figure}

Given the initial annotation budget $n_0$, we should split the numbers to each subset. We define the original number of each subset $s$ as $n_{s}$ and the selected number to label as $n^l_{s}$.  The number of samples in different categories often varies, and samples from minority categories (such as Night-Rainy and Overtake) are typically challenging and critical, requiring more attention. Therefore, we introduce a parameter $\gamma$ to control the proportions of each subset $P_{s}$. The proportion calculation of first-level weather-lighting subset is specified as follows:
\begin{equation}
\label{eq:px}
    \begin{aligned}
         P_{x} = \frac{n^{\gamma}_{x}}{\sum_{z \in \{\text{DS, DR, NS, NR} \}} n^{\gamma}_{z}} , 
        \text{ where } x \in \{\text{DS, DR, NS, NR}\}. 
    \end{aligned}
\end{equation}

The annotation number for each subset $s$ is $n^l_s = n_0 P_s$. 
When $\gamma = 1$, it indicates an absolute uniform distribution, where each category is chosen equally. If $\gamma < 1$, it signifies a bias towards categories with fewer total samples. For the second-level subset consisting of four driving scenarios, the process is similar:
\begin{equation}
\label{eq:pxy}
    \begin{aligned}
         & P_{x, y} = P_x \times \frac{n^{\gamma}_{x, y}}{\sum_{z \in \{\text{L, R, O, S} \}} n^{\gamma}_{x, z}} , \\ 
        \text{where } x  \in \{ & \text{DS, DR, NS, NR}\} \text{ and } y \in  \{\text{L, R, O, S} \}.
    \end{aligned}
\end{equation}

 Fig.~\ref{fig:egodiversity} gives the detailed intuitive selection of the initial selecting process based on a multi-way tree. Firstly, the entire dataset is divided into four first-level subsets DS, DR, NS and NR based on weather and lighting conditions. Secondly, within each of these subset, further divisions are made based on driving commands, resulting in four second-level subsets L, R, O and S from each weather-lighting subset. Finally, based on the available sample budget $n_{x,y} = n_0 P_{x,y}$ in each second-level subset, a selection is made at regular intervals within the sorted speeds.

\subsection{Criterion Design for Incremental Selection}
\label{sec:increSelection}
In this section, we introduce how we incrementally annotate a new portion of clips based on the model trained with those already annotated ones. We will use the intermediate model to conduct inference on the unannotated clips and the subsequent selection is based on these outputs. Still, we take a planning-oriented view 
 and we introduce three criteria for subsequent data selection: Displacement Error, Soft Collision, and Agent Uncertainty. 

\noindent\textbf{Criterion I: Displacement Error (DE).} Denote $\mathcal{L}_{DE}$ as the distance between the predicted planning route $\tau$ of the model and the human trajectory $\tau^*$ recorded in the dataset.
\begin{equation}
    \mathcal{L}_{DE} = \frac{1}{T} \sum_{t = 1}^{T} \|\tau_t - \tau^*_t\|_2,
\end{equation}
where $T$ represents the frames in the scenes. Since the Displacement Error itself is a performance metric (without the need of annotation), it naturally becomes the first and most crucial criterion in active selection. 

\noindent\textbf{Criterion II: Soft Collision (SC).} Define $\mathcal{L}_{SC}$ as the distance between the predicted ego-trajectory and predicted agent-trajectory. Similar to \cite{jiang2023vad}, we will filter out low-confidence agent predictions by a threshold $\epsilon_{a}$. In each scene, we select the shortest distance as a measure of the danger coefficient. Simultaneously, we maintain a positive correlation between the term and the closest distance:
\begin{equation}
    \mathcal{L}_{SC} = \sum_{t=1}^{T} \exp \left(- \min_{a \in \text{agents}} (\tau_{t, ego} - \tau_{t, a})\right).
\end{equation}

We use Soft Collision as one criterion because: On one hand, different from the Displacement Error, the calculation of Collision Ratio depends on the annotation of 3D bounding boxes for objects, which are not available in unlabeled data. Thus, we should be able to calculate the criterion solely based on the inference results of the model. On the other hand, consider a Hard Collision criterion: if the predicted ego trajectory would have a collision with other predicted agents' trajectories, we assign it as 1 and otherwise as 0. However, it could cause too few samples with label 1 since the collision rate of state-of-the-art models in AD are usually very small (less than $1\%$). Hence, we choose to use the closest distance to other objects as a substitute for the Collision Rate metric. When the distance to other vehicles or pedestrians is too close, the risk is considered to be significantly higher. In short, Soft Collision serves as an effective indicator to measure the likelihood of a collision, which could provide dense supervisions.

\noindent\textbf{Criterion III: Agent Uncertainty (AU).} 
The prediction of surrounding agents' future trajectories naturally has uncertainty~\cite{ide-net} and thus the motion prediction module usually generates multiple modalities and corresponding confidence scores. We aim to select those data where nearby agents has high uncertainties. Specifically, we filter out faraway agents by a distance threshold $\delta_d$ and calculate the weighted entropy of the predict probabilities of multiple modalities of remaining agents. Suppose
the number of the modalities is $N_m$ and the confidence scores of a agent under different modalities are $\mathcal{P}_{i}(a)$ where $i \in \{1, ..., N_m\}$. Then, the Agent Uncertainty $\mathcal{L}_{AU}$ can be defined as :
\begin{equation}
    \begin{aligned}
        \mathcal{L}_{AU}  = \sum_{a \in \text{agent}} \mathcal{W}(a) \mathcal{H}(a) 
        = -\sum_{a \in \text{agent}} \exp(\delta_d - d_a) \left(\sum_{i=1}^{N_m} \mathcal{P}_i(a) \log \mathcal{P}_i(a)  \right),
    \end{aligned}
\end{equation}
where $d_a$ is the predicted distance between the agent and ego vehicle, $\mathcal{W}$ represents the weight and $\mathcal{H}$ is the entropy.

\noindent\textbf{Overall Loss} The loss of active selection is defined as:
\begin{equation}
    \mathcal{L} = \mathcal{L}_{DE} + \alpha \mathcal{L}_{SC} + \beta \mathcal{L}_{AU},
    \label{eq:overall_loss}
\end{equation}
where $\alpha, \beta$ are hyper-parameters. We select the top $n_i$ unannotated clips with the largest overall loss, where $n_i$ denotes the number of clips that can be annotated in iteration $i$.


\subsection{Overall Active Learning Paradigm}
\label{sec:pipeline}
In summary, Alg.~\ref{alg:pipeline} presents the entire workflow of our method. Given the available budget $B$, the initial selection size $n_0$, the number of active selections made at each step $n_i$, and $M$ total selections stage. We start by initializing the selection using randomization or the Ego-Diversity method described in Sec.~\ref{sec:initSelection}.
Then, we train the network using the current annotated data. 
Based on the trained network, we make predictions on the unlabeled pool and calculate the overall loss described in Sec~\ref{sec:increSelection}. Finally, we sort the samples based on the overall loss and select the top $n_i$ samples to be annotated in the current iteration. We repeat this process until the iterations reach the upper bound $M$ and the selected number of samples reaches the upper limit $B$.

\begin{algorithm}[tb!]
    \caption{\textbf{Pseudo-code for ActiveAD}}
    \label{alg:pipeline}
    \KwInput{Unlabeled pool $\mathcal{P}^u = \{X_i\}_{i\in[N]}$, labeled pool $\mathcal{P}^l = \emptyset$, model $f(\cdot; w)$, annotation budget $B$, initial number $n_0$, active selection iterations $M$, selection number per iteration $n_{itr}$, original numbers of each subset $n_{x}$ and $n_{x,y}$, hyper-parameters $\alpha, \beta, \gamma$.}
    
    \vspace{5pt}
    Initialize annotation dataset indexes $\mathcal{K} = \emptyset$.
    \vspace{5pt}
    
    \If {\text{Using Ego-Diversity based initialization}} {
        \vspace{5pt}
        
       \For{First-level subset $x$ in \{DS, DR, NS, NR\}}{
            \vspace{5pt}
            Calculate first-level proportion  $P_{x} = n^{\gamma}_{x} / \sum_{z} n^{\gamma}_{z}$ where $z \in \{\text{DS, DR, NS, NR} \}$ by Eq.~\ref{eq:px}.
            \vspace{5pt}
            
            \For{Second-level subset $y$ in \{L, R, O, S\}}{
            Calculate second-level proportion $P_{x, y} = P_x \times n^{\gamma}_{x, y} / \sum_{z} n^{\gamma}_{x, z}$ where $z \in \{\text{L, R, O, S} \}$ by Eq.~\ref{eq:pxy}. \vspace{5pt}

            Set the annotation number $n^l_{x,y} = n_0 P_{x,y}$.
            \vspace{5pt}

            Sort the subset $x,y$ according to the speed in ascending order and select $n^l_{x,y}$ indexes at regular intervals, then add them to $\mathcal{K}$.
            }
       }
    }
    \Else {
        Randomly select $n_0$ samples $\mathcal{K} = \{k_i \in [N]\}_{i \in [n_0]}$.
    }
    

    
    
    \For{$itr \in \{1,2,\dots,M\}$}{
        \vspace{5pt}
        Update $\mathcal{P}^u_{\mathcal{K}} = \mathcal{P}^u - \{X_{k_i}\}_{i \in [\sum_{j=0}^{itr-1} n_j]}$ and $\mathcal{P}^l_\mathcal{K} = \{(X_{k_i}, Y_{k_i})\}_{i \in [\sum_{j=0}^{itr-1} n_j]}$.
        \vspace{5pt}
    
        Train the model $f(\cdot; w)$ from scratch using $\mathcal{P}^l_{\mathcal{K}}$.
        \vspace{5pt}
        
        Inference  on unlabeled pool $\mathcal{P}^u_{\mathcal{K}}$ to calculate Loss $\mathcal{L} = \mathcal{L}_{DE} + \alpha \mathcal{L}_{SC} + \beta \mathcal{L}_{AU}$ in Eq.~\ref{eq:overall_loss} for each sample. 
        \vspace{5pt}

        Sort the samples in the descending order of $\mathcal{L}$.
        \vspace{5pt}

        Select the former $n_{itr}$ indexes and add them to the $\mathcal{K}$ so that $\mathcal{K} = \{k_i \in [N]\}_{i \in [\sum_{j=0}^{itr} n_j]}$.
    }
    
    \KwOutput{Return annotation indexes $\mathcal{K} = \{k_i \in [N]\}_{i \in [B]}$}
\end{algorithm}

\begin{table*}[tb!]
    \centering
    \caption{\textbf{Planning Performance.} ActiveAD outperforms general active learning baseline in all annotation budget settings. Moreover, ActiveAD with $30\%$ data achieves even slightly better planning performance than using the entire dataset for training. VAD with $^*$ indicates that we have updated the results, which are better than those reported in the original works. UniAD with $\dag$ indicates that we have employed the metrics from VAD to update the results (Refer to Appendix \ref{app:evalutaion_metric} for more details).}
    \vspace{-5pt}
    \resizebox{\textwidth}{!}
    {
      \begin{tabular}{l|c|c|cccc||cccc}
      \toprule[1.2pt]
       \multirow{2}{*}{\textbf{Base Model}} &
       \multirow{2}{*}{\textbf{Percent}} &
       \multirow{2}{*}{\textbf{Selection Method}} &
       \multicolumn{4}{c||}{\textbf{Average L2 (m) } $\downarrow$} & \multicolumn{4}{c}{\textbf{Average Collision (\%) } $\downarrow$ }   \\ 
        & & & 1s & 2s & 3s & Avg. & 1s & 2s & 3s & Avg.    \\ 
       \midrule[1.0pt]
        ST-P3~\cite{hu2022st} & 100\% & - & 1.33 & 2.11 & 2.90 & 2.11 & 0.23 & 0.62 & 1.27 & 0.71 \\

        UniAD$^{\dag}$~\cite{hu2023planning} & 100\% & - & 0.42 & 0.64 & 0.91 & 0.67 & - & - & - & - \\
        
        VAD-Base$^*$~\cite{jiang2023vad} & 100\% & - & 0.39 & 0.66 & 1.01 & 0.69 & 0.08 & 0.16 & 0.37 & 0.20\\
        VAD-Tiny$^*$~\cite{jiang2023vad} & 100\% & - & 0.38 & 0.68 & 1.04 & 0.70 & 0.15 & 0.22  & 0.39 & 0.25 \\
        \midrule[0.6pt]
        \multirow{3}{*}{VAD-Tiny} & 10\% & Random & 0.51 & 0.83 & 1.23 & 0.86 & 0.40 & 0.62 & 0.98 & 0.67  \\
         & 10\% & ActiveFT~\cite{xie2023active} & 0.54 & 0.88 & 1.29  & 0.90 & 0.20 & 0.41 & 0.81 & 0.47 \\
         & 10\% & \textbf{ActiveAD(Ours)} & \textbf{0.47} & \textbf{0.80}  & \textbf{1.21}  & \textbf{0.83} & \textbf{0.13} & \textbf{0.35}  & \textbf{0.80} &  \textbf{0.43} \\
         
        \midrule[0.6pt]
        \multirow{5}{*}{VAD-Tiny} & 20\% & Random & 0.49 & 0.80 & 1.17 & 0.82 & 0.36 & 0.49 & 0.77 & 0.54 \\
         & 20\% & Coreset~\cite{sener2018active} & 0.48 & 0.78 & 1.16 & 0.81 & 0.20 & 0.40 & 0.69 & 0.43   \\
         & 20\% & VAAL~\cite{sinha2019variational} & 0.54 & 0.89 & 1.31 & 0.91 & 0.17 & 0.38 & 0.66 & 0.40  \\
         & 20\% & ActiveFT~\cite{xie2023active}  & 0.50 & 0.82 & 1.21 & 0.84 & 0.27 & 0.42 & 0.63 & 0.44  \\
         & 20\% & \textbf{ActiveAD(Ours)} & \textbf{0.44} & \textbf{0.73} & \textbf{1.10} & \textbf{0.76} & \textbf{0.18} & \textbf{0.36} & \textbf{0.62} & \textbf{0.39}\\
        
        \midrule[0.6pt]
        \multirow{5}{*}{VAD-Tiny} & 30\% & Random & 0.45 & 0.76 & 1.12 & 0.78 & 0.17 & 0.30 & 0.63 & 0.37 \\
         & 30\% & Coreset~\cite{sener2018active} & 0.43 & 0.71 & 1.06 & 0.73 & 0.43 & 0.51 & 0.68 & 0.54   \\
         & 30\% & VAAL~\cite{sinha2019variational} & 0.46 & 0.79 & 1.19 & 0.81 & 0.18 & 0.33 & 0.54  & 0.35  \\
         & 30\% & ActiveFT~\cite{xie2023active}  & 0.46 & 0.76 & 1.13 & 0.78 & 0.18 & 0.35 & 0.63  & 0.39 \\
         & 30\% & \textbf{ActiveAD(Ours)} & \textbf{0.41} & \textbf{0.66} & \textbf{0.97} & \textbf{0.68} & \textbf{0.10} & \textbf{0.18} & \textbf{0.36} & \textbf{0.21}\\

        \bottomrule[1.2pt]
        \end{tabular}
    }  
    \vspace{-10pt}
    \label{tab:planning_result}
\end{table*}

\section{Experiments}

We conduct experiments on the widely used nuScenes dataset~\cite{caesar2020nuscenes} in line with the peer works~\cite{hu2023planning}. All experiments are implemented using PyTorch and run on RTX 3090 and A100 GPUs. Source code will be made publicly available.

\subsection{Experimental Setup}
\label{sec:exp_setup}

\textbf{Dataset \& Metrics.} 
The nuScenes~\cite{caesar2020nuscenes} dataset consists of 1,000 scenes, each lasting 20 seconds. It provides comprehensive annotations, including 3D bounding boxes for 23 classes and 8 attributes. The scenes are captured by 6 cameras, providing a 360 degree horizontal FOV, and the keyframes are annotated at a frequency of 2Hz. It covers a wide range of locations, time, and weather conditions. In line with previous works~\cite{hu2022st, hu2022goal, jiang2023vad}, we evaluate the planning performance using the Displacement Error (L2 loss) and Collision Rate metrics.

\textbf{End-to-end AD Models.}
We selected latest works ST-P3~\cite{hu2022st}, UniAD~\cite{hu2022goal} and VAD~\cite{jiang2023vad} as our baseline models. Among them, the latest VAD demonstrates superior planning performance. Moreover, it achieves substantial reductions in computational overhead, and accelerates the training. Therefore, we adopt the lightweight version, VAD-Tiny, as the base model for subsequent experiments. We also include VAD-Based results in Sec.~\ref{app:vad_base} of the supplementary materials.

\textbf{Active Learning Baselines.}
As mentioned in  Sec.~\ref{sec:activelearning}, end-to-end autonomous driving is a novel and under-explored task for active learning. Directly transferring existing active learning methods, which are typically based on predictive probability analysis, is nontrivial. In particular, we select three methods as baselines that are relatively more transferable and relevant to this task: Coreset~\cite{sener2018active}: a feature selection-based approach; VAAL~\cite{sinha2019variational}: a task-agnostic method; and ActiveFT~\cite{xie2023active}, which utilizes pre-trained features. Coreset utilizes the embeddings  prior to the trajectory planning head~\cite{jiang2023vad} as the input features. VAAL and ActiveFT take the raw images as inputs. The former employs an adversarial learning paradigm to discriminate unlabeled samples, while the latter uses ResNet50~\cite{he2016deep} as the pretrained model for feature extraction, which is also adopted as the default backbone network in VAD~\cite{jiang2023vad}. ActiveFT selects all data within the budget at once, eliminating the need for iterative selection.

\textbf{Implementation Details.} 
We set the chosen budget $B$ as $30\%$ of the data volume: initially selecting $10\%$ in the data pool, followed by an additional $10\%$ in each subsequent selection round, for a total of two selection rounds. In each round, the model is retrained and used for the next round selection. We apply VAD-Tiny as the base model using the default hyper-parameter configuration. The confidence threshold $\epsilon_a$ and distance threshold $\delta_d$ is set to $0.5$ and $3.0$m respectively. 
For the initial selection, we set driving scenario threshold $\tau_c=4$ and diversity partitioning parameter $\gamma=0.5$.
For the overall loss in Eq.~\ref{eq:overall_loss}, we normalize the criteria $\mathcal{L}_{DE}, \mathcal{L}_{SC}, \mathcal{L}_{AU}$ to $[0,1]$ according to all scenes value respectively and set hyper-parameters $\alpha=1$ and $\beta=1$.
We use AdamW~\cite{loshchilov2017decoupled} optimizer and Cosine Annealing~\cite{loshchilov2016sgdr} scheduler to train VAD-Tiny $20$ epochs with weight decay of $0.01$ and initial learning rate of $2 \times 10^{-4}$.

\subsection{Performance by Planning Metrics}
In Tab.~\ref{tab:planning_result} , we present the performance of all active learning models when choosing $10\%$, $20\%$, $30\%$ of training samples. In the supplementary material, we further give results of 40\%, 50\% and we observe that the performance is saturated at 30\%, which again demonstrates the long-tail nature of AD data. We observe that traditional Active Learning methods perform poorly, lacking any significant advantage over random selection. In contrast, ActiveAD demonstrates significant advantages across the three different granularity ratios for data selection, highlighting the effectiveness of our method. This design enables improved sample selection and annotation for end-to-end planning-oriented autonomous driving. This is particularly relevant because manual annotation of samples for autonomous driving is resource-intensive and time-consuming. An astonishing finding is that \textbf{ActiveAD achieves comparable or even better performance by utilizing a carefully selected $30\%$ of the data compared to training with the entire $100\%$ dataset}. We believe that this finding is both intriguing and significant as it challenges the notion that more data necessarily leads to better performance. Current methods often focus on refining model structures while overlooking the importance of judicious data utilization. We argue that the data we select is more representative and informative, enabling to eliminate unnecessary noise and trivial samples that may cause adverse effects.




\subsection{Ablation Study}

\begin{table*}[tb!]
    \centering
    \caption{\textbf{Ablation for Designs.} ``RA" and ``ED" indicate the Random and Ego-Diversity based initial set selection. ``DE", ``SC" and ``AU" indicates Displacement Error, Soft Collision and Agent Uncertainty, respectively. All combinations with ``ED" utilize the same $10\%$ of data for initialization. The criteria $\mathcal{L}_{DE}, \mathcal{L}_{SC}, \mathcal{L}_{AU}$ are normalized to $[0,1]$ respectively and we set hyperparameters $\alpha$ and $\beta$ as $1$.}\vspace{-5pt}
    \resizebox{\textwidth}{!}{
    \begin{tabular}{c|cc|ccc||ccc||ccc}
      \toprule[1.0pt]
       \multirow{2}{*}{\textbf{ID}} &
       \multicolumn{2}{c|}{\textbf{Initiation}} &
       \multicolumn{3}{c||}{\textbf{Active Selection}} &
       \multicolumn{3}{c||}{\textbf{Average L2 (m) }} & \multicolumn{3}{c}{\textbf{Average Collision (\%)}}   \\ 
        & \ RA & ED & \  DE & \quad SC & AU & $10\%$ & $20\%$ & $30\%$ & $10\%$ & $20\%$ & $30\%$    \\ 
       \midrule[0.8pt]
        1 & \checkmark & - & - & - & - & 0.86  & 0.82 & 0.78 & 0.67  & 0.54 & 0.37 \\
        2 & - & \checkmark & - & - & - & 0.83 \minus{0.03} & 0.78 \minus{0.04} & 0.74 \minus{0.04} & 0.41 \minus{0.26} & 0.40 \minus{0.14} & 0.34 \minus{0.03}\\
        3 & - & \checkmark & \checkmark & - & - & 0.83 \minus{0.03} & \textbf{0.68 \minus{0.14}} & 0.70 \minus{0.08} & 0.41 \minus{0.26} & 0.39 \minus{0.15} & 0.35 \minus{0.02}\\
        4 & - & \checkmark & \checkmark & \checkmark & - & 0.83 \minus{0.03} & 0.81 \minus{0.01} & 0.73 \minus{0.05} & 0.41 \minus{0.26} & \textbf{0.35 \minus{0.19}} & 0.26 \minus{0.11} \\
        5 & \checkmark & - & \checkmark & \checkmark & \checkmark & 0.86 \minus{0.00} & 0.80 \minus{0.02} & 0.71 \minus{0.07} &  0.67 \minus{0.00} & 0.38 \minus {0.16} & 0.26 \minus {0.11} \\
        6 & - & \checkmark & \checkmark & \checkmark & \checkmark & 0.83 \minus{0.03}  & 0.76 \minus{0.06} & \textbf{0.68 \minus{0.10}}  & 0.41 \minus{0.26} & 0.39 \minus{0.15} & \textbf{0.21 \minus{0.16}} \\
      \bottomrule[1.0pt]
    \end{tabular}
    }
    \label{tab:metric_influ}
\end{table*}

\noindent \textbf{Effectiveness of Designs.}
Tab.~\ref{tab:metric_influ} shows the contributions of all the proposed components described in Sec.~\ref{sec:active} to the final planning performance, including Displacement Error (L2) and Collision Rate. Our proposed Ego-Diversity based method exhibits superior performance in initial $10\%$ data selection, particularly in reducing the collision rate from $0.67\%$ to $0.41\%$, thus providing a better initialization for subsequent model training.

During the subsequent active selection process, different metrics focus on different aspects. For instance, Displacement Error (DE) emphasizes the disparity between predicted and ground truth trajectories, effectively reducing the L2 loss of driving when solely utilized. However, it is regrettable that the performance of Collision Rate is unsatisfactory. Meanwhile, even with an increase in data volume, the results obtained using $30\%$ of the data can be worse than those achieved with $20\%$ of the data in terms of L2 performance. Indeed, when solely focusing on a single metric, it is easy to overlook other valuable information, which can potentially lead to overfitting.

Moreover, we believe that avoiding collisions requires considering information from surrounding vehicles. Relying solely on Displacement Error makes it challenging to optimize the selection process. Therefore, the inclusion of Soft Collision metric can improve the performance in this aspect. In the case of selecting $30\%$ of the data, the collision rate decreased significantly from $0.35\%$ to $0.26\%$, demonstrating a notable reduction. Additionally, considering the various possibilities of different objects in different environments, leveraging Agent Uncertainty can enhance the selection of complex scenarios. Agent Uncertainty assists in better optimizing both two planning metrics when the data volume increases. By incorporating these designs, ActiveAD has achieved outstanding performance. We also demonstrate that only  utilizing our incremental selection based on random initialization results in significant performance improvements.

\begin{table}[tb!]
    \renewcommand{\arraystretch}{1.1}
    \centering
    \caption{\textbf{Ablation for Ego-Diversity Hyperparameter.} We enumerated the distributions obtained by selecting $10\%$ data under various $\gamma$ and compared their performance. \# represents the numbers of scene occurrence.  
    }
    \vspace{-5pt}
    \resizebox{0.8\linewidth}{!}{%
    \begin{tabular}{l|cccc|cccc|cc}
        \toprule
            \textbf{Diversity} & \multicolumn{4}{c|}{\textbf{Weather-Lighting}} & \multicolumn{4}{c|}{\textbf{ Driving-Command }} & \multicolumn{2}{c}{\textbf{Metric}}    \\ 
            \cline{2-11} 
            \textbf{Parameter} & \#DS &  \#DR & \#NS & \#NR &  \#S \quad & \#R \quad & \#L  & \#O & L2 (m) $\downarrow$ & CR ($\%$) $\downarrow$ \\
            \hline
            Complete & 491 & 125 & 71 & 13 & 423 & 132 & 112 & 33 & 0.70 & 0.25 \\
            \hline
            $\gamma=1$ & 49 & 12 & 7 & 2 & 40 & 13 & 11 & 6 & 0.90 & 0.46  \\
            $\gamma=0.8$ & 43 & 14 & 10 & 3 & 35 & 14 & 14 & 7 & 0.88 & \textbf{0.41} \\
            $\gamma=0.5$ & 34 & 17 & 13 & 6 & 27 & 17 & 16 & 10 & \textbf{0.83} & 0.43 \\

            \bottomrule
    \end{tabular}%
    } 
    \vspace{-10pt}
    \label{tab:ego_diversity_hyperparameter}
\end{table}

\noindent\textbf{Ego-Diversity Hyperparameter Analysis.}
We introduce the hyperparameter $\gamma$ in Sec.~\ref{sec:initSelection} to adjust the proportion of initial selection based on the number of samples. Whether in real scenarios or for model training purposes, these corner cases with fewer samples are often challenging and require special attention. Therefore, we choose to increase the focus on minority classes for the case of $\gamma \leq 1$. Tab.~\ref{tab:ego_diversity_hyperparameter} displays the results of our preliminary experiments with different parameter values. We observe that when $\gamma = 1$, it ensures the stability of the selection process and provides velocity-based uniform selection compared to random selection. $\gamma = 0.8$ exhibits better performance in Collision Rate, while $\gamma = 0.5$ shows a clear advantage in Displacement Error (L2). Considering that the impact of Collision Rate diminishes when L2 is large, we select $\gamma = 0.5$ as the fixed parameter for subsequent model training and selection. Additionally, we did not extensively tune other parameters, such as $\alpha,\beta,\epsilon_a, \tau_c$, as their default values described in Sec.~\ref{sec:exp_setup} already yielded satisfactory results.

\begin{figure*}[tb!]
    \centering     \includegraphics[width=\linewidth]{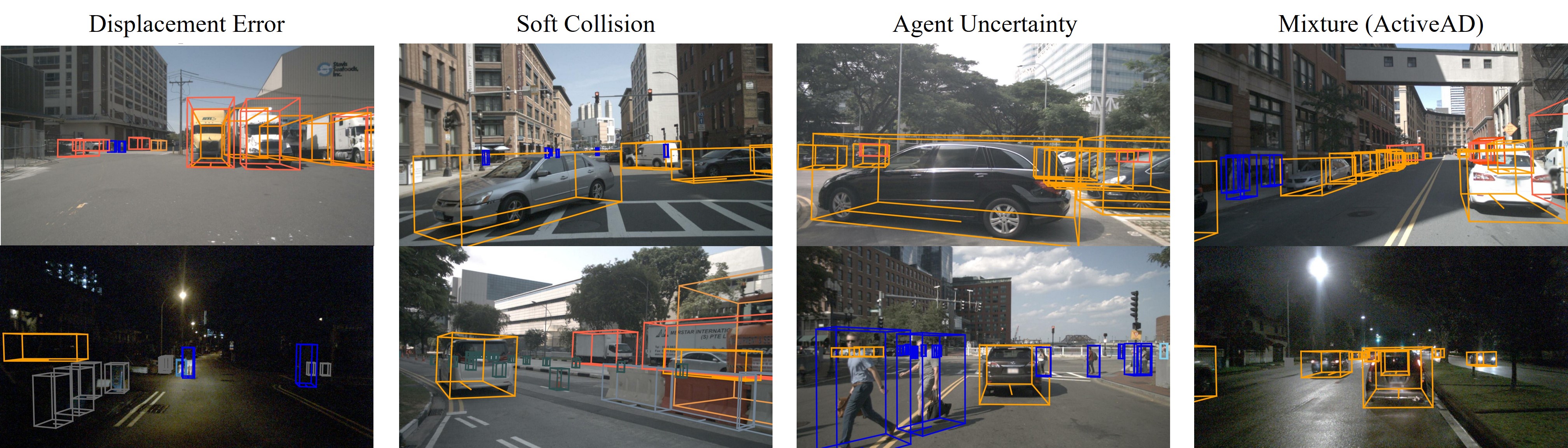}
    \vspace{-14pt}
    \caption{\textbf{Selected Scenes Visualization}. Front camera images selected according to the criterion of Displacement Error (col 1), Soft Collision (col 2), Agent Uncertainty (col 3) and Mixture (col 4)  based on the model trained on $10\%$ data. Mixture represents our final selection strategy ActiveAD, with considerations for the previous three scenarios. }  
    \label{fig:scene_visualization}
\end{figure*}

\begin{table*}[tb!]
    \centering
    \caption{\textbf{Performance under Various Scenarios.}  Average L2 (m) / Average Collision Rate ($\%$) of active models using $30\%$ data under various weather / lighting and driving-command conditions. The smaller the better performance.}\vspace{-5pt}
    \resizebox{\textwidth}{!}
    {
      \begin{tabular}{l|cccc||cccc||c}
      \toprule[1.2pt]
        
       \multirow{2}{*}{\textbf{Method}} &
       \multicolumn{4}{c||}{\textbf{Weather / Lighting} } & \multicolumn{4}{c||}{\textbf{Driving-Command}  } &  \\ 
        & Day & Night & Sunny & Rainy & Go Straight & Turn Left & Turn Right & Overtake  & All \\ 
        \midrule[0.8pt]
        Complete & 0.67 / 0.27 & 1.01 / 0.14 & 0.70 / 0.32 & 0.72 / 0.04 & 0.69 / 0.32 & 0.74 / 0.13 & 0.67 / 0.20 &  0.84 / 0.13 & 0.70 / 0.25 \\
        \midrule[0.6pt]
        Random & 0.72 / 0.26 & 1.29 / 1.25 & 0.78 / 0.39 & 0.79 / 0.26 & 0.70 / 0.22 &  0.89 / 1.03 & 0.86 / 0.32 & 1.05 / 0.22 & 0.78 / 0.37 \\
        Coreset~\cite{sener2018active} & 0.71 / 0.57 & 0.97 / 0.27 &  0.72 / 0.65 & 0.78 / 0.06 & 0.69 / 0.67 & 0.78 / 0.31 & 0.78 / 0.38 & 0.96 / 0.14 & 0.73 / 0.54  \\
        VAAL~\cite{sinha2019variational} & 0.78 / 0.34 & 1.09 / 0.34 & 0.80 / 0.40 & 0.89 / 0.12 & 0.79 / 0.38 & 0.86 / 0.34 & 0.82 / \textbf{0.20} & 0.96 / 0.18 & 0.81 / 0.35  \\
        ActiveFT~\cite{xie2023active}  & 0.76 / 0.37 & 1.08 / 0.43 & 0.79 / 0.40  & 0.78 / 0.28 & 0.70 / 0.35 & 0.88 / 0.62 & 0.91 / \textbf{0.20} & 1.18 / 0.44 & 0.79 / 0.38 \\
        \textbf{ActiveAD(Ours)} & \textbf{0.64 / 0.20} & \textbf{1.03 / 0.31} & \textbf{0.68 / 0.24} & \textbf{0.68 / 0.07} & \textbf{0.62 / 0.21} & \textbf{0.74 / 0.25} & \textbf{0.80 / 0.20} & \textbf{0.85 / 0.13} & \textbf{0.68 / 0.21}\\

        \bottomrule[1.2pt]
        \end{tabular}
    }  
    \label{tab:Condition_result}
    \vspace{-10pt}
\end{table*}

\noindent\textbf{Various Scenarios Analysis.}
We study the performance of the active methods under diverse scenarios. Tab.~\ref{tab:Condition_result} demonstrates that our method, ActiveAD, outperforms competitors in all cases, highlighting its superiority. ActiveAD exhibits strong robustness and excels in challenging situations, including rainy or nighttime conditions, as well as during overtaking maneuvers known for their higher difficulty. Furthermore, we achieve comparable performance while using $30\%$ available data, as opposed to utilizing the entire dataset for comparison.



\begin{figure}[tb!]
    \centering     
    \includegraphics[width=0.9\linewidth]{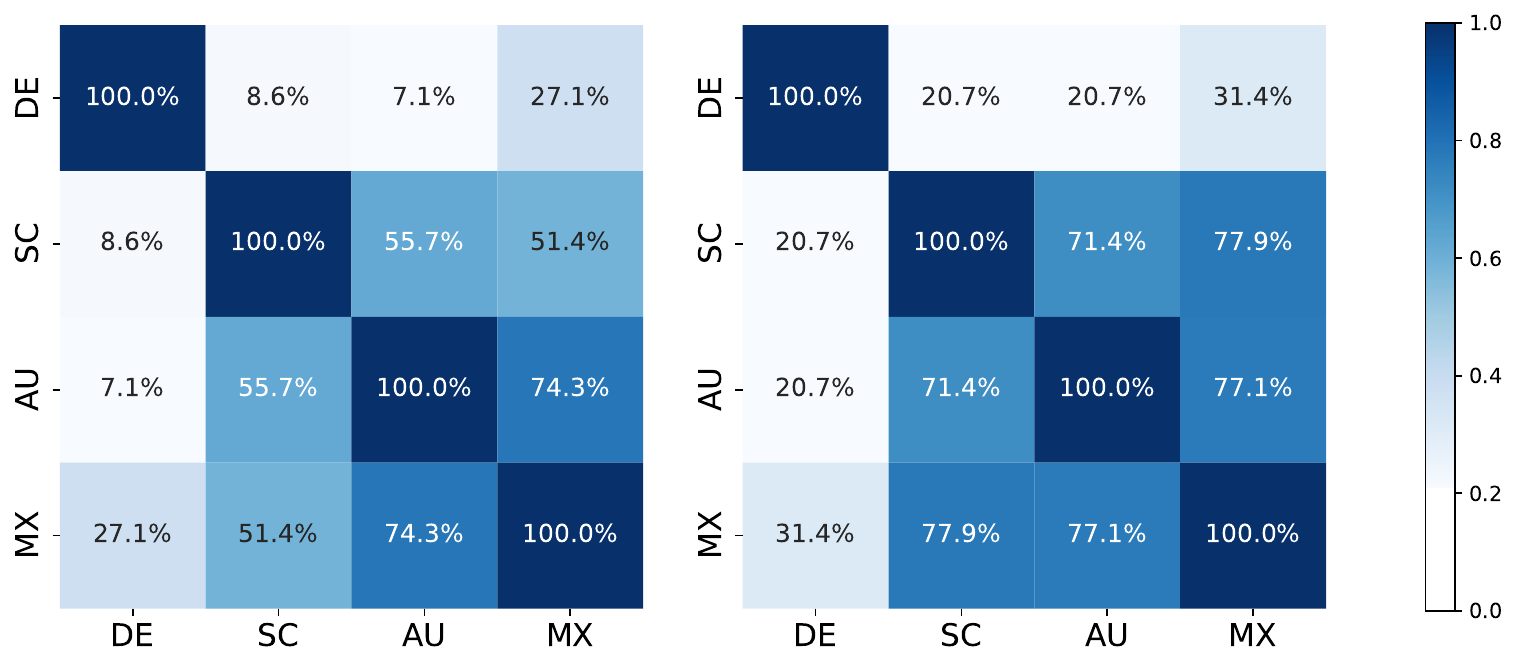}
    \vspace{-5pt}
    \caption{\textbf{Similarity between multiple criteria}. It shows the repetition rate of the $10\%$ (Left) and $20\%$ (Right) new sampled scenes selected by four criteria: Displacement Error (DE), Soft Collision (SC), Agent Uncertainty (AU) and Mixture (MX).}  
    \label{fig:Similarity}
    \vspace{-10pt}
\end{figure}

\noindent\textbf{Selected Scenes Visualization.}
Based on the model trained on $10\%$ of the data, Fig.~\ref{fig:scene_visualization} illustrates the selection of representative scenarios using different metrics. The scenarios selected based on Displacement Error include complex maneuver trajectories such as lane changes and pedestrian avoidance. The scenarios selected based on Soft Collision often involve situations where the ego vehicle is in close proximity to other vehicles or obstacles, posing a risk. Examples include waiting at intersections for other vehicles to make turns, dense traffic in adjacent lanes, or situations with a high concentration of surrounding obstacles. Agent Uncertainty focuses on challenging road conditions, such as flickering lights, overtaking behaviors, vehicle reversing, and pedestrians crossing. ActiveAD combines considerations from all three criteria to select comprehensive samples across various scenarios. Fig.~\ref{fig:Similarity} illustrates the overlap rate among the scenes selected based on these different criteria. In comparison, ActiveAD with mixture criterion demonstrates a better coverage of scenarios considered by individual criteria and emphasizes more on truly complex situations to enhance data quality for achieving excellent model performance.

\section{Conclusion}
In addressing the high cost and long-tail issues of data annotation for end-to-end autonomous driving, we are the first to develop a tailored active learning scheme ActiveAD. ActiveAD introduces novel task-specific diversity and uncertainty metrics based on the planning-oriented philosophy. Extensive experiments demonstrate the effectiveness of our approach, surpassing  general peer methods by a significant margin and achieving comparable performance to the state-of-the-art model, using only $30\%$ of the data. This represents a meaningful exploration of end-to-end autonomous driving from a data-centric perspective and hope our work can inspire future research and discoveries.

\vspace{2mm}
\noindent\textbf{Limitations}: 
As a planning-oriented active learning approach, we have achieved significant effects within the planning metrics. However, the 30\% of data we selected still falls far short in training the model's perception and prediction capabilities compared to using 100\%. Experiments in Appendix~\ref{app:per_pre_performance} demonstrate that model perception and prediction gradually strengthen with an increase in data volume. This is typical in similar fields, such as active learning for segmentation, and our method has not overcome this bottleneck. Nonetheless, within the E2E-AD framework, we have effectively identified valuable samples, reduced annotation costs, and avoided overfitting.

\clearpage  

%
%
\bibliographystyle{splncs04}
\bibliography{main}

\clearpage

\appendix

\section{Experiments Details}
\label{app:exp_detail}
\subsection{Experiments Setup}
\textbf{End-to-end Autonomous Driving Models.} 
ST-P3~\cite{hu2022st} is an interpretable end-to-end vision-based network for autonomous driving, achieving better spatial-temporal feature learning. UniAD~\cite{hu2022goal} leverages the information from multiple preceding tasks to enhance goal-oriented planning and demonstrates outstanding performance in all aspects including perception, prediction and planning. VAD~\cite{jiang2023vad} introduces a vectorized paradigm as a substitute for the dense rasterized scene representation employed in previous studies. This approach facilitates a more concentrated analysis of instance-level structural information, leading to excellent end-to-end planning performance. Moreover, it achieves substantial reductions in computational requirements, decreases the reliance on training devices, and accelerates the training speed. Consequently, we adopt the lightweight version, VAD-Tiny, as the start point for our experiments. 

\textbf{Active Learning Baselines.}
As mentioned in  Sec~\ref{sec:activelearning}, end-to-end autonomous driving is actually a novel and underexplored task for active learning. It is difficult to directly transfer existing active learning approaches which are usually based on predictive probability analysis into this task. Therefore, we choose three classic methods that are more transferable and relevant for the task as baselines: Coreset, a feature selection-based approach; VAAL, a task-agnostic method; and ActiveFT, which utilizes pre-trained features.  
1) Coreset~\cite{sener2018active} formulates the data selection process as a k-Center problem on the learned embeddings of both labeled and unlabeled data. We utilize the features prior to the trajectory planning head~\cite{jiang2023vad} as the embeddings. 2) VAAL~\cite{sinha2019variational} utilizes the adversarial learning paradigm,  
employing a variational autoencoder (VAE)~\cite{kingma2013auto} to extract the image features in the nuscenes dataset, along with a discriminator network that distinguishes between labeled and unlabeled images. The VAE aims to deceive the discriminator by making it classify all samples as labeled data, while the discriminator strives to accurately identify the unlabeled samples in the data pool. Based on this approach, the selected unlabeled samples are then annotated by annotators. 3) ActiveFT~\cite{xie2023active} utilizes pretrained features to optimizes the distance between the distributions of labeled and unlabeled sets. In state-of-the-art autonomous driving methods, BEV feature~\cite{li2022bevsurvey} is the commonly used representation. We adopt ActiveFT to use BEV features to select data, and its strength lies in the ability to select all data under the budget at once without the need for iterative selection.

\textbf{Annotation Budget}. 
In the scenario of active learning, the annotation budget is typically predetermined. Considering the complexity of end-to-end autonomous driving models and the diversity of tasks (including the final planning task as well as auxiliary perception and prediction tasks), we have set the annotation budget as 30\%. Meanwhile, We further report the performance of ActiveAD with the budget from 10\% to 50\% of the data in Tab.~\ref{tab:perception_prediciton_performance}. We observe that the planning performance is saturated around 30 \% and thus we choose 30\% as the stop threshold in the main paper.

\begin{table}[t]
\caption{\textbf{All tasks' performance under different selection ratio.}}
\vspace{-5pt}
\centering
\resizebox{1\linewidth}{!}{%
    \begin{tabular}{c|cc||ccccccc||cccc}
    \toprule
    \multirow{2}{*}{\textbf{Ratio}} & \multicolumn{2}{c||}{\textbf{Planning}} & \multicolumn{7}{c||}{\textbf{Perception}} & \multicolumn{4}{c}{\textbf{Prediction}} \\
     & Avg. L2 $\downarrow$   & Avg. Col. $\downarrow$  & NDS $\uparrow$ & mAP $\uparrow$ & mATE $\downarrow$ & mASE $\downarrow$ & mAOE $\downarrow$ & mAVE $\downarrow$ & mAAE $\downarrow$ & minADE $\downarrow$ & minFDE $\downarrow$ & MR  $\downarrow$  & EPA $\uparrow$\\ 
    \hline
    10\%  & 0.83 & 0.43 & 16.56 & 9.80 & 0.95 & 0.43 & 0.98 & 1.31 & 0.47 & 1.28 & 1.89 & 0.195 & 0.230 \\
    20\%  & 0.76 & 0.39 & 21.46 & 14.77 & 0.83 & 0.45 & 0.84 & 0.99 & 0.49 & 1.10 & 1.59 & 0.161 & 0.373\\
    30\%  & 0.68 & \textbf{0.21} & 25.60 & 15.85 & 0.84 & 0.39 & 0.78 & 0.83 & 0.40 & 1.01 & 1.43  & 0.147 & 0.402\\
    40\%  & \textbf{0.66} & 0.24 & 27.12 & 18.20 & 0.81 & 0.36 & 0.83 & 0.79 & 0.35 & 0.96 & 1.36  & 0.145 &  0.414 \\
    50\%  & 0.68 & 0.23 & 29.29 & 19.72 & 0.85 & 0.34 & 0.80 & 0.76 & 0.31 & 0.93 & 1.28 & 0.142 & 0.430\\
    100\% & 0.70 & 0.25 & \textbf{36.11} & \textbf{26.65} & \textbf{0.74} & \textbf{0.31}  & \textbf{0.76} & \textbf{0.67} & \textbf{0.23} & \textbf{0.84} & \textbf{1.16} & \textbf{0.134} & \textbf{0.534}  \\ 
    \bottomrule
    \end{tabular}
    }
    \label{tab:perception_prediciton_performance}
    \vspace{-10pt}
\end{table}

\subsection{Metrics Explanation}
\label{app:evalutaion_metric}

In this paper, we utilize the evaluation metrics from VAD~\cite{jiang2023vad}, which is consistent with ST-P3~\cite{hu2022st}. Therefore, the results from these two papers can be directly applied. Recently, inconsistencies in the UniAD metrics~\cite{hu2023planning} have been identified within the community~\cite{mao2023language,li2023ego}. We reference the content in \cite{mao2023language} to provide more details about the evaluation metrics. 
The output trajectory $\tau$ is formatted as 6 waypoints in a 3-second horizon, i.e., $\tau=$ $\left[\left(x_1, y_1\right),\left(x_2, y_2\right), ..., \left(x_6, y_6\right)\right]$.
Then, the L2 loss is computed as:
$$
l_2=\sqrt{(\tau-\hat{\tau})^2}=\left[\sqrt{\left(x_i-\hat{x}_i\right)^2+\left(y_i-\hat{y}_i\right)^2}\right]_{i=1}^6,
$$
where $l_2 \in \mathbb{R}^{6 \times 1}$ and $\hat{\tau}$ denotes ground truth trajectory. Then, the average L2 loss $\bar{l}_2 \in \mathbb{R}^{6 \times 1}$ can be computed by averaging $l_2$ for each sample in the test set.

UniAD~\cite{hu2023planning} uses the value in the exact timestep as the L2 loss at the $k$-th second $(k=1,2,3)$ :
$$
L_{2, k}^{\mathrm{UniAD}}=\bar{l}_2[2 k] .
$$

ST-P3~\cite{hu2022st} and VAD~\cite{jiang2023vad} use the the average error from 0 to $k$ second as L2 loss at the $k$-th second:
$$
L_{2, k}^{\mathrm{VAD}}=\frac{\sum_{t=1}^{2 k} \bar{l}_2[t]}{2 k} .
$$

Given the collision times $\mathcal{C} \in \mathbb{N}^{6 \times 1}$ at each timestep. Similarly, UniAD reports the collision $\mathcal{C}_k^{\text {uniad }}$ at the $k$-th second $(k=1,2,3)$ as $\mathcal{C}[2 k]$, while VAD reports $\mathcal{C}_k^{\text {VAD }}$ as the average from 0 to $k$ second.

Besides the variations in calculation methodologies, there is a distinction in the generation of ground truth occupancy maps between the two metrics. UniAD exclusively accounts for the vehicle category in creating ground truth occupancy maps, whereas ST-P3 and VAD incorporates both vehicle and pedestrian categories. This discrepancy results in different collision rates for the same planned trajectories when evaluated by these metrics, although it has no effect on the L2 error measurement. As a result, the collision rate in UniAD may be higher than reported, and this has been confirmed in \cite{li2023ego} where VAD demonstrates superior performance in terms of collision rates. Consequently, we use a '-' in Tab.\ref{tab:planning_result} instead of displaying specific values.

Taking into account the advantages of VAD in terms of model lightweighting (for instance, the ability to train using a 3090 GPU) as well as its leading position in comprehensive performance, we explore active learning based on the VAD model in this paper. This exploration is conducted from the perspective of data, aiming to provide insightful analysis.

\subsection{Experiment Results for VAD-Base}
\label{app:vad_base}
Tab.~\ref{tab:app_planning_result} presents the experimental results of our method based on the VAD-Base model.  Compared to the baseline of random selection, our method—whether it be the one-time sample selection based on Ego-Diversity or the complete method that performs Incremental Selection starting from an initial dataset—has shown significant advantages. Consistent with the conclusions in the main paper, using 30\% of the data, our approach achieves performance on par with using the entire dataset, validating the effectiveness and universality of our method.

\begin{table*}[tb!]
    \centering
    \caption{\textbf{Planning Performance with VAD-Base.} ActiveAD (w/o incremental) refers to the selection of all data solely based on diversity selection. ActiveAD (w/ incremental) indicates performing incremental selection based on an initial set.}
    \vspace{-5pt}
    \resizebox{\textwidth}{!}
    {
      \begin{tabular}{l|c|c|cccc||cccc}
      \toprule[1.2pt]
       \multirow{2}{*}{\textbf{Base Model}} &
       \multirow{2}{*}{\textbf{Percent}} &
       \multirow{2}{*}{\textbf{Selection Method}} &
       \multicolumn{4}{c||}{\textbf{Average L2 (m) } $\downarrow$} & \multicolumn{4}{c}{\textbf{Average Collision (\%) } $\downarrow$ }   \\ 
        & & & 1s & 2s & 3s & Avg. & 1s & 2s & 3s &  Avg. \\ 
       \midrule[1.0pt]
       ST-P3~\cite{hu2022st} & 100\% & - & 1.33 & 2.11 & 2.90 & 2.11 & 0.23 & 0.62 & 1.27 & 0.71 \\

        UniAD$^{\dag}$~\cite{hu2023planning} & 100\% & - & 0.42 & 0.64 & 0.91 & 0.67 & - & - & - & - \\
        VAD-Base$^*$~\cite{jiang2023vad} & 100\% & - & 0.39 & 0.66 & 1.01 & 0.69 & 0.08 & 0.16 & 0.37 & 0.20\\
        VAD-Tiny$^*$~\cite{jiang2023vad} & 100\% & - & 0.38 & 0.68 & 1.04 & 0.70 & 0.15 & 0.22  & 0.39 & 0.25 \\
        
        \midrule[0.6pt]
        \multirow{2}{*}{VAD-Base} 
         & 10\% & Random &  0.49 & 0.81 & 1.20 & 0.83 & 0.38 & 0.57 & 0.91 & 0.62   \\
         & 10\% & ActiveAD(w/o incremental) & \textbf{0.48} & \textbf{0.76} & \textbf{1.14} & \textbf{0.79} & \textbf{0.24} & \textbf{0.43} & \textbf{0.68} & \textbf{0.45} \\

        \midrule[0.6pt]
        \multirow{3}{*}{VAD-Base} 
         & 20\% & Random & 0.47 & 0.78 & 1.15 & 0.80 & 0.32 & 0.47 & 0.75 & 0.51  \\
         & 20\% & ActiveAD(w/o incremental) & 0.44 & 0.75 & 1.10 & 0.76 & 0.25 & \textbf{0.34} & \textbf{0.61} & 0.40\\
         & 20\% & ActiveAD(w/ incremental) & \textbf{0.42} & \textbf{0.70} & \textbf{1.08} & \textbf{0.73} & \textbf{0.16} & 0.35 & 0.64 & \textbf{0.38} \\

        \midrule[0.6pt]
        \multirow{3}{*}{VAD-Base} 
         & 30\% & Random & 0.44 & 0.74 & 1.08 & 0.75 & 0.16 & 0.34 & 0.54 & 0.35  \\ 
         & 30\% & ActiveAD(w/o incremental) & 0.42 & 0.71 & 1.05 & 0.73 & 0.14 & 0.29 & 0.49 & 0.31\\
         & 30\% & ActiveAD(w/ incremental) & \textbf{0.40} & \textbf{0.67} & \textbf{0.93} & \textbf{0.67} & \textbf{0.09} & \textbf{0.21} & \textbf{0.35} & \textbf{0.22} \\
               
        \bottomrule[1.2pt]
        \end{tabular}
    }  
    \vspace{-15pt}
    \label{tab:app_planning_result}
\end{table*}


\section{Perception and Prediction Performance.}
\label{app:per_pre_performance}
Existing end-to-end training models~\cite{hu2022goal,jiang2023vad} often utilize visual information as auxiliary tasks to assist core objective planning. The main experiment shown in Tab.~\ref{tab:planning_result}, demonstrates our advantage in planning metrics, while we are also curious about perception and prediction task performance. Tab.~\ref{tab:perception_prediciton_performance} displays the performance after training with different proportions of data. The perception metrics include NDS(nuScenes detection score), mAP(mean Average Precision), mATE(mean Average Translation Error), mASE(mean Average Scale Error), mAOE(mean Average Orientation Error), mAVE(mean Average Velocity Error), mAAE(mean Average Attribute Error) which are sourced from the nuScenes dataset setting~\cite{caesar2020nuscenes}. The prediction metrics include minADE (minimum Average Displacement Error), minFDE (minimum Final Displacement Error) and MR (Miss Rate) and EPA (End-to-end Prediction Accuracy)~\cite{hu2023planning}.

We have observed that there still exists a significant performance gap in these metrics between utilizing a small amount of data and using complete data. This observation aligns with common sense in active learning tasks~\cite{sener2018active,sinha2019variational, xie2023active, zhan2022comparative}, where a small sample size  can not outperform the entire dataset in traditional image classification and segmentation tasks. This raises the question of how to balance other losses in end-to-end autonomous driving, considering planning as the ultimate objective, and whether there are better training paradigms. Our active learning approach provides a means to optimize training data while reducing costs. We believe that future work on multitask learning or hard case mining holds promise for enhancing planning performance.

\end{document}